# Leveraging Natural Language Processing to Unravel the Mystery of Life: A Review of NLP Approaches in Genomics, Transcriptomics, and Proteomics.

Ella Rannon[1], David Burstein[1,2]

[1]The Shmunis School of Biomedicine and Cancer Research, George S. Wise Faculty of Life Sciences, Tel-Aviv University, Tel-Aviv, Israel

[2]Correspondence: davidbur@tauex.tau.ac.il

## Abstract

Natural Language Processing (NLP) has transformed various fields beyond linguistics by applying techniques originally developed for human language to the analysis of biological sequences. This review explores the application of NLP methods to biological sequence data, focusing on genomics, transcriptomics, and proteomics. We examine how various NLP methods, from classic approaches like word2vec to advanced models employing transformers and hyena operators, are being adapted to analyze DNA, RNA, protein sequences, and entire genomes. The review also examines tokenization strategies and model architectures, evaluating their strengths, limitations, and suitability for different biological tasks. We further cover recent advances in NLP applications for biological data, such as structure prediction, gene expression, and evolutionary analysis, highlighting the potential of these methods for extracting meaningful insights from large-scale genomic data. As language models continue to advance, their integration into bioinformatics holds immense promise for advancing our understanding of biological processes in all domains of life.

## Introduction

Natural language processing (NLP) is a rapidly evolving field in computer science, focusing on the automated analysis of text. Its primary goal is enabling machines to understand natural languages, extract information accurately, and generate valuable insights[1]. Language models (LMs) have become central to numerous NLP tasks, such as sentiment analysis, machine translation, question answering, information retrieval, summarization, and more[1–6]. LMs analyze human language patterns by learning the probability distribution of basic language units (i.e., words, sub-words, or characters), which are termed tokens.

In recent years, significant advancements in NLP have been driven by breakthroughs in deep learning. These advances encompass innovative neural network architectures, novel training approaches, increased computing power, and access to large text corpora. Notably, transformer-based LMs like BERT and GPT-3 have achieved unprecedented success in the field of NLP by leveraging self-supervised learning on massive unlabeled datasets[7,8]. Models such as BERT, trained using the masked language modeling (MLM) task, learn contextual vector embeddings of words and sentences, which are extremely useful for downstream tasks. Meanwhile, autoregressive language models, like the Generative Pre-trained Transformer (GPT) models[8–10], are trained on the next token prediction task and have made impressive advancements in natural language generation.

The impact of NLP has extended far beyond its traditional domains, leading to significant applications in various fields. These techniques, originally designed to enable computers to understand human language, are now being successfully applied to the "languages" of biology, including DNA, RNA, and protein sequences[11–15]. This approach holds immense promise for decoding the biological syntax and semantics, potentially revolutionizing our understanding of genomics and computational biology.

In this review, we survey studies and models applying NLP approaches to analyze biological sequence data. We examine various NLP techniques, ranging from classic approaches like word2vec to cutting-edge models employing transformers and hyena operators and their application to genomics, transcriptomics, and proteomics. We analyze these models based on their architectural design and the type of biological input they consider as a sentence, be it DNA, RNA, protein sequences, or entire genomes. Additionally, we discuss the biological tasks these algorithms aim to address, while also considering their limitations and potential pitfalls.

## Classical Embedding Methods

The field of distributed word representations gained significant momentum with the introduction of word2vec. Word2vec employs a shallow neural network to transform one-hot word vectors — sparse vectors the size of the vocabulary, where only the index corresponding to the target word is set to one while all others are zero — into dense, distributed word vectors of continuous values. Word2vec employs two types of model architecture to produce these representations: skip-gram and continuous bag-of-words (CBOW). The skip-gram model uses a given center word in a sentence to predict surrounding words, while the CBOW model predicts the center word based on a window of context words Figure 1A). The weight matrix trained using these approaches serves as the representation of words in the vocabulary[16]. An extension of word2vec, known as doc2vec, was developed to capture paragraph-level context and provide features for each sentence[17]. FastText, a more advanced embedding method based on the skip-gram model, incorporates sub-word information in its representation learning process, allowing it to better handle words not present in the training data (out of vocabulary)[18].

Global Vectors for Word Representation (GloVe) presents an alternative embedding approach that utilizes word co-occurrence matrices. GloVe offers advantages in terms of training efficiency and scalability to large corpora[19]. Despite the significant advancements in representation learning facilitated by word2vec, fastText, and GloVe, they do not account for word order, meaning sentences composed of the same words in different orders yield the same vector representations. Moreover, they cannot express the semantic polysemy of words, as they produce a single vector for each token, failing to capture multiple meanings of words in different contexts. Nevertheless, these embedding techniques have laid the groundwork for more advanced NLP models, setting the stage for contextual embeddings that address these limitations.

## LSTMs

Long Short-Term Memory (LSTM) networks are a type of recurrent neural network (RNN) that process sequences sequentially. They learn representations that capture information from a given position and all its preceding positions. In these models, the probability of the $t^{th}$ word in a sentence is calculated using the hidden state of the word $t$, which depends on the previous words in the sentence and their corresponding hidden states. During training, the model's weights are updated to maximize the joint probability for all words in the sentence. Since the hidden state is computed recursively based on word order, LSTM-based models facilitate context-aware learning[20].

To incorporate information from both preceding and succeeding tokens, many LSTM-based language models utilize bidirectional LSTMs (BiLSTMs). These models combine two separate LSTMs operating in forward and backward directions within each layer[21] (Figure 1B). This bidirectional approach allows the model to capture context from both sides of each word in a sentence. A notable implementation of this concept is the Embeddings from Language Models (ELMo) approach, which leverages the hidden states of a multi-layer BiLSTM to create powerful, context-aware word representations[22]. ELMo thus addresses the issue of polysemy by providing context-specific embeddings based on the entire input sentence.

While RNN-based models can theoretically learn long-term dependencies, in practice, the hidden state of a particular word tends to depend more heavily on nearby words, with the influence of distant words often diminishing or converging to zero. Additionally, RNNs struggle with large-scale learning due to limited parallelization, as they sequentially process all past states. They also suffer from the vanishing gradient problem, which can reduce training efficiency[23]. LSTM models address the vanishing gradient issue more effectively by introducing memory cells alongside hidden states[24]. These memory cells store long-term information and are updated at each step through gating mechanisms. The gates determine what information is retained from the previous memory cell and what is updated based on the current hidden state[20]. To better overcome the inherent limitations of RNNs, the transformer-based architecture was developed. Transformers do not rely on past hidden states to capture the dependency on the previous words. Instead, they process entire sentences simultaneously, enabling parallel computation and mitigating the vanishing gradient issue and performance degradation associated with long-term dependencies.

### Transformers

Transformers revolutionized NLP by introducing a novel approach to learning representations through explicit calculation of attention vectors across all positions in a sequence. This self-attention mechanism learns the representation for each position while "attending to" every other position in the same sequence, thus helping to address the polysemy issue. This mechanism quantifies directly and in parallel the dependencies between words, even those distant in the linear sequence. While computationally demanding, transformers overcome most limitations of previous models, making them powerful tools for various NLP tasks[25].

The original transformer architecture comprised an encoder and a decoder unit (Figure 1C). The encoder, consisting of multiple identical blocks of attention and feed-forward layers, encodes the input sequence into a vector space. The encoder's output is passed to the decoder, which decodes the intermediate sequence and outputs the desired output sequence. The decoder includes an additional attention sub-layer between the original two sub-layers to perform multi-head attention between the encoder stack's output and the output of the decoder's self-attention sub-layer[25].

A prominent transformer architecture, that uses only the encoder portion, is the Bidirectional Encoder Representations from Transformers (BERT). BERT consists of multiple transformer layers and is pre-trained using two tasks: Masked Language Modeling (MLM) and Next Sentence Prediction (NSP)[7]. In MLM, the model predicts original tokens of randomly masked words based on their context, enabling it to jointly consider the context before and after each word. NSP involves binary classification to determine if two input sentences are semantically consecutive. The outputs from BERT's pre-trained layers serve as distributed representations of input sentences, which can be fine-tuned for various downstream tasks such as question answering, sentiment analysis, or named entity recognition.

Transformers have also proven highly effective as autoregressive language models[8–10] and achieved state-of-the-art performance in various tasks, including machine translation, question answering, and text classification[26]. Their ability to capture long-range dependencies and process entire sequences in parallel has made them a cornerstone of modern NLP research and applications. However, the transformers' quadratic time and space complexity limit the length of sequences they can effectively process.

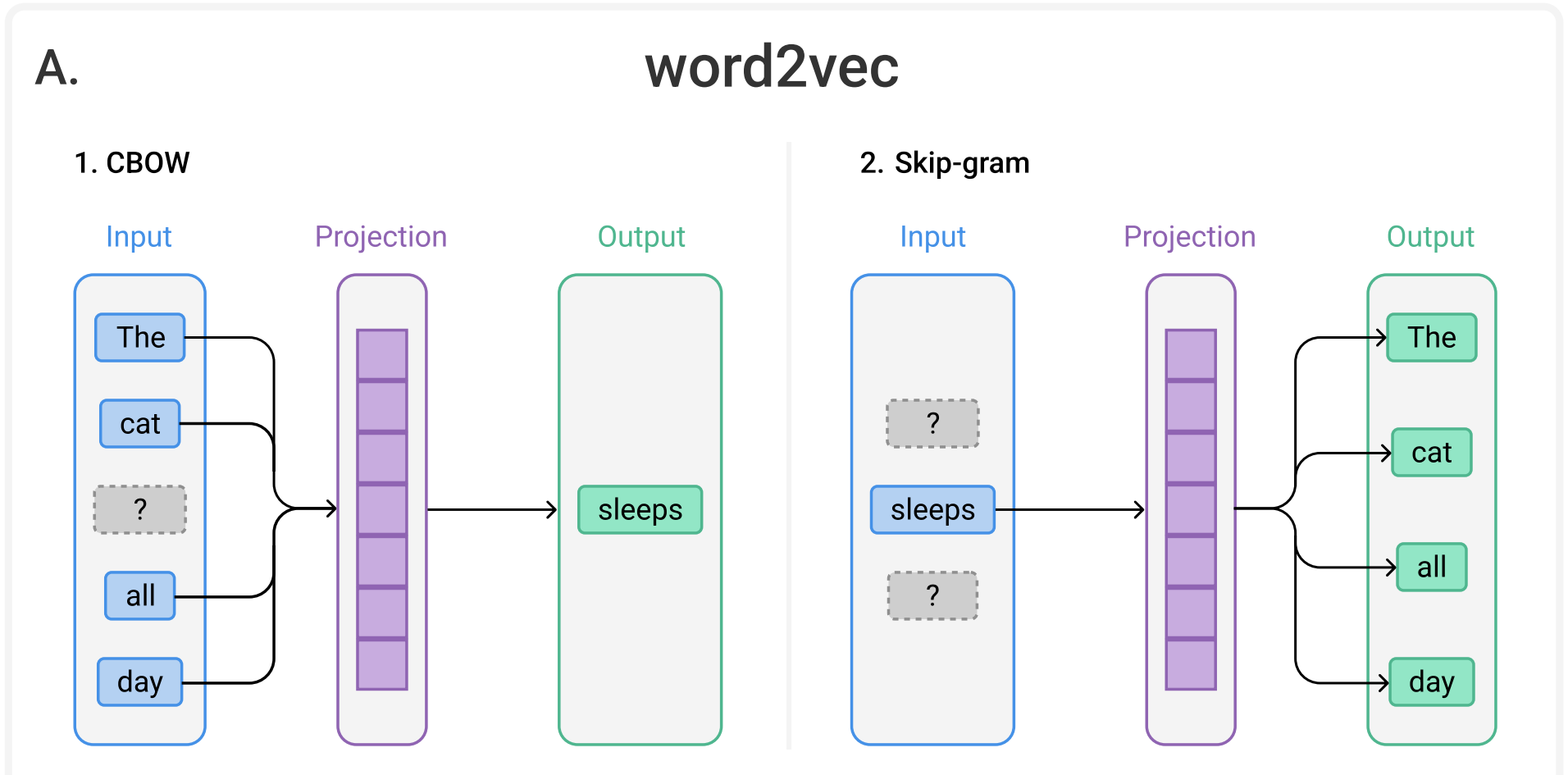

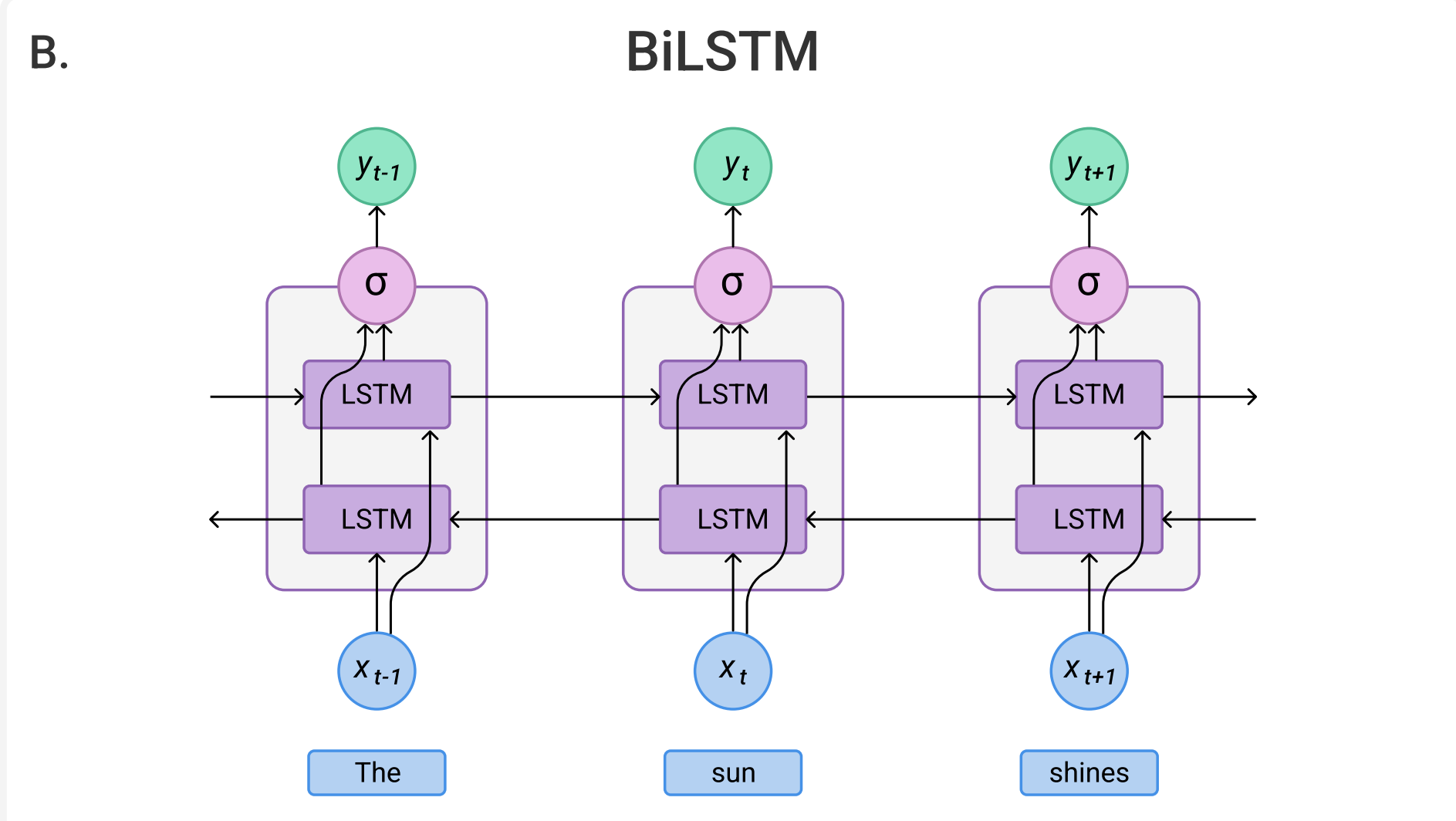

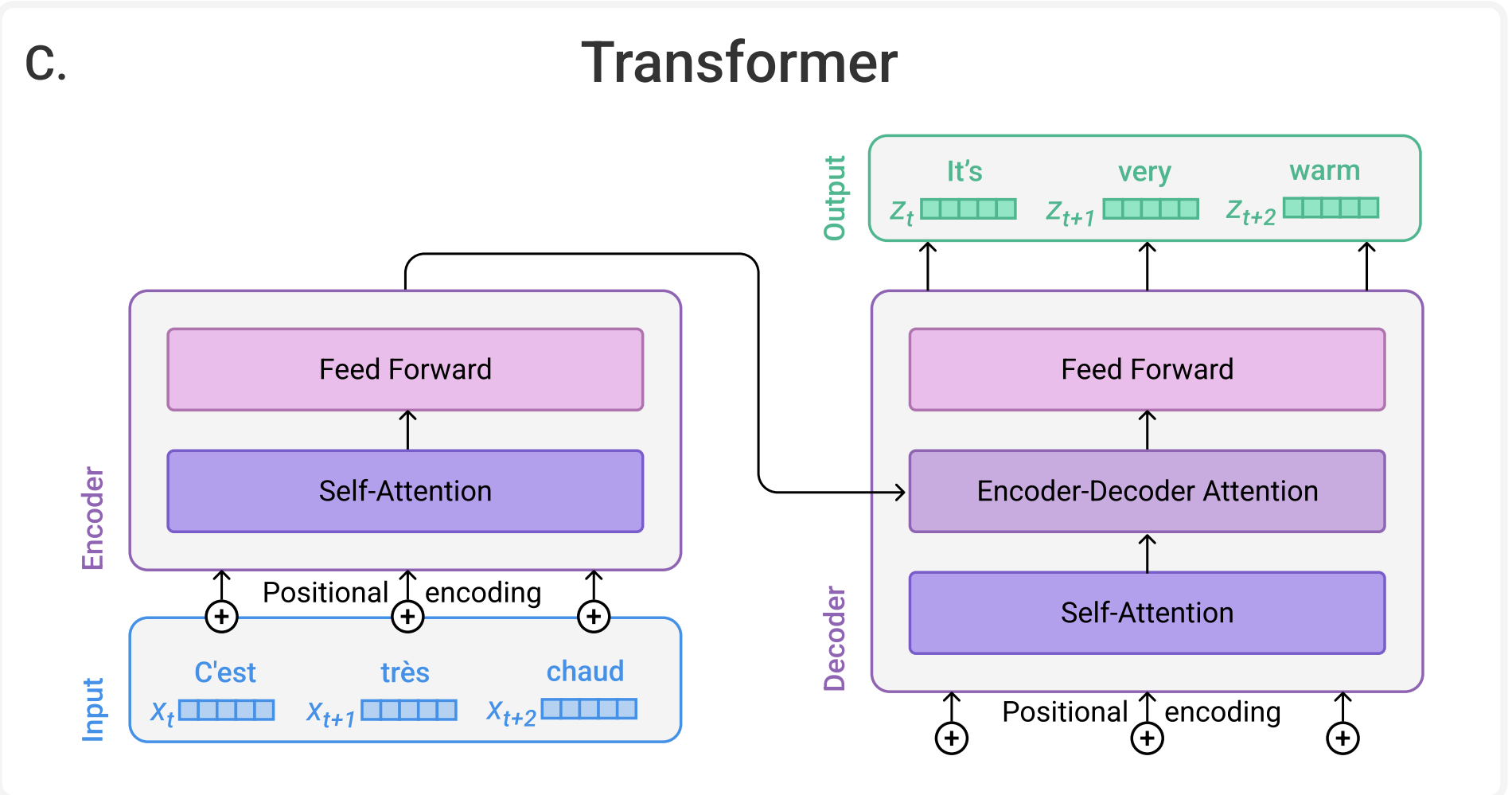

**Figure 1. Overview of key NLP architectures. A.** The two word2vec architectures. Left: Continuous Bag-of-Words (CBOW), where the center word is predicted from a window of context words; Right: Skip-gram, where context words are predicted based on the center word. **B.** Bidirectional LSTM (BiLSTM) architecture: The probability of a word is computed using its hidden state, which incorporates information from both preceding and succeeding words in the sentence, along with their respective hidden states. **C.** Transformer architecture: Comprising an encoder stack and a decoder stack, each input token is represented by a learned embedding and positional encoding. Encoders contain self-attention layers to capture dependencies between tokens and feed-forward. The final encoder's attention matrices are integrated into the decoder stack via the encoder-decoder attention layers. Decoders iteratively process the sequence, predicting the next token, which is appended to the input sequence for subsequent iterations.

## Hyena Architecture

The Hyena architecture represents a recent and significant advancement in language modeling, offering a sub-quadratic time complexity as an alternative to "traditional" attention mechanisms. It leverages a class of data-controlled operators consisting of recurrent multiplicative gating interactions and long convolutions. This design allows for efficient evaluation in sub-quadratic time[27], and efficiency enables Hyena to perform in-context learning on sequences of considerable length, addressing one of the key limitations of traditional transformer architectures.

Empirical evaluations have demonstrated that Hyena achieves comparable performance to attention-based mechanisms while substantially reducing computational time complexity[27]. This improvement in efficiency facilitates the processing of more extended contexts, potentially broadening the scope of applications for large language models. Nevertheless, it is noteworthy that while the Hyena architecture shows promise, further research is needed to fully assess its capabilities and limitations across various NLP tasks and domains.

## NLP in Genomics, Transcriptomics, and Proteomics

Natural language is composed of characters, with meaning constructed through grammar and semantics. Similarly, biological sequences can be viewed as "sentences" composed of distinct letters (e.g., nucleotides, amino acids, or genes), governed by biophysical, biochemical, and evolutionary rules that determine their "meanings", such as function and structure[28]. This analogy highlights that the meaning of biological sequences, like natural language, is defined by characters in a specific order (Figure 2A).

A fundamental concept in NLP is learning the semantics of a sentence through the context of its words, based on the hypothesis that tokens occurring in similar contexts tend to have similar meanings. Analogously, research has shown that the genomic context, i.e., the set of genes located near a given gene, provides crucial insights into the gene's function[29–31]. This parallel between natural language and biological sequences has driven bioinformatics research to adopt NLP methods, aiming to deepen our understanding of how functions and structures are embedded in biological sequences[11–14,32].

Moreover, NLP addresses a significant challenge in genomics, transcriptomics, and proteomics: the vast data accumulation due to advancements in high-throughput sequencing technologies. While these technologies generate a large amount of data, they do not inherently provide biological insights or interpretation[33]. NLP methods, which require only sequence data as input, offer a promising approach to extract meaningful insights from this wealth of information.

Language models have emerged as a powerful tool for distilling information from extensive protein sequence databases. From sequence data alone, these models uncover evolutionary, structural, and functional organization across protein space. These models encode amino-acid sequences into distributed vector representations that capture their structural and functional properties, as well as assess the evolutionary fitness of sequence variants[34–38]. These language models capture complex dependencies between amino acids and can be trained on diverse sequences rather than being focused on individual families. Moreover, transformer-based language models have also been successfully applied to tasks such as processing DNA sequences[39–41], gene context[42], gene expression prediction[43], and other bioinformatic applications[14] (Figure 2B).

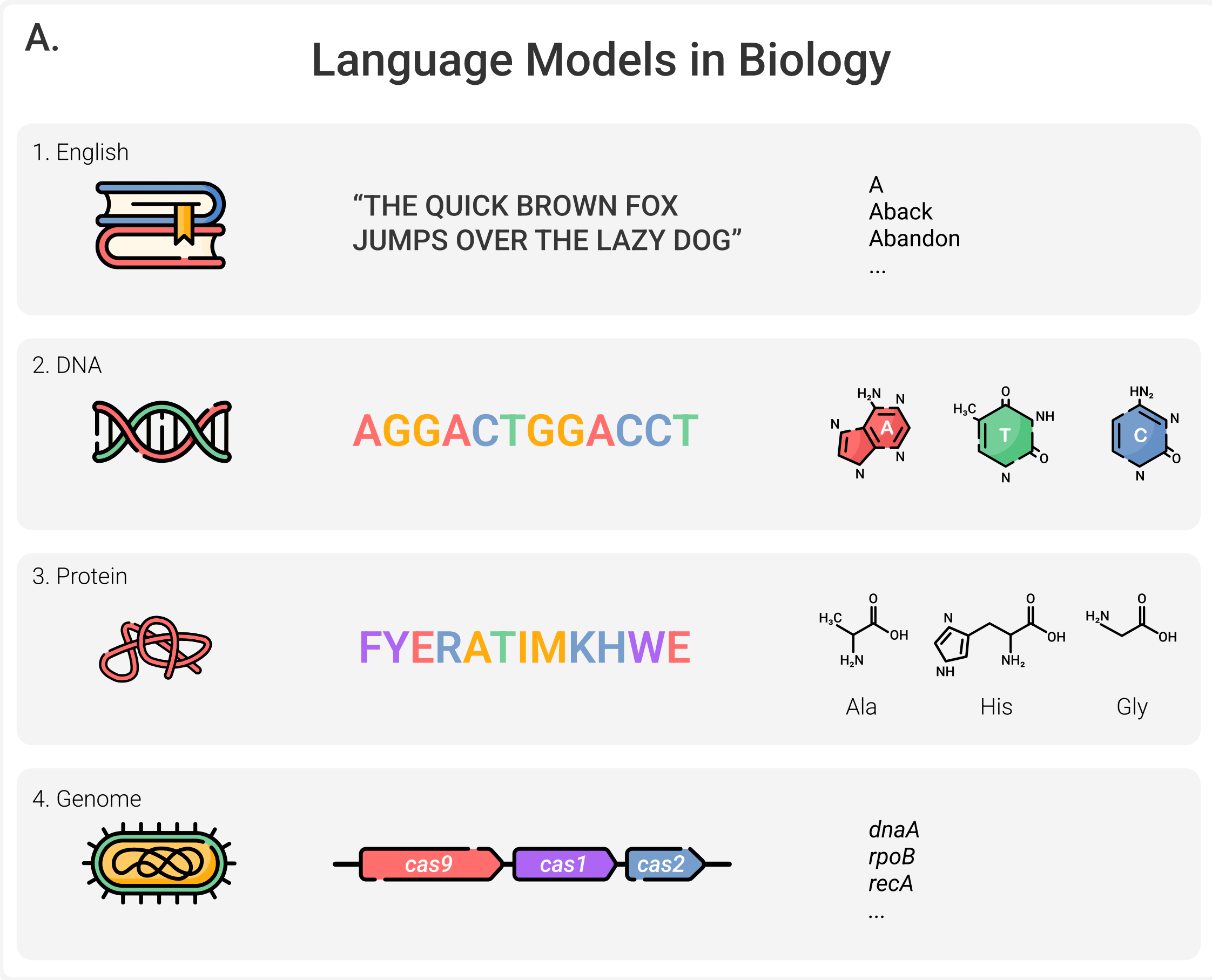

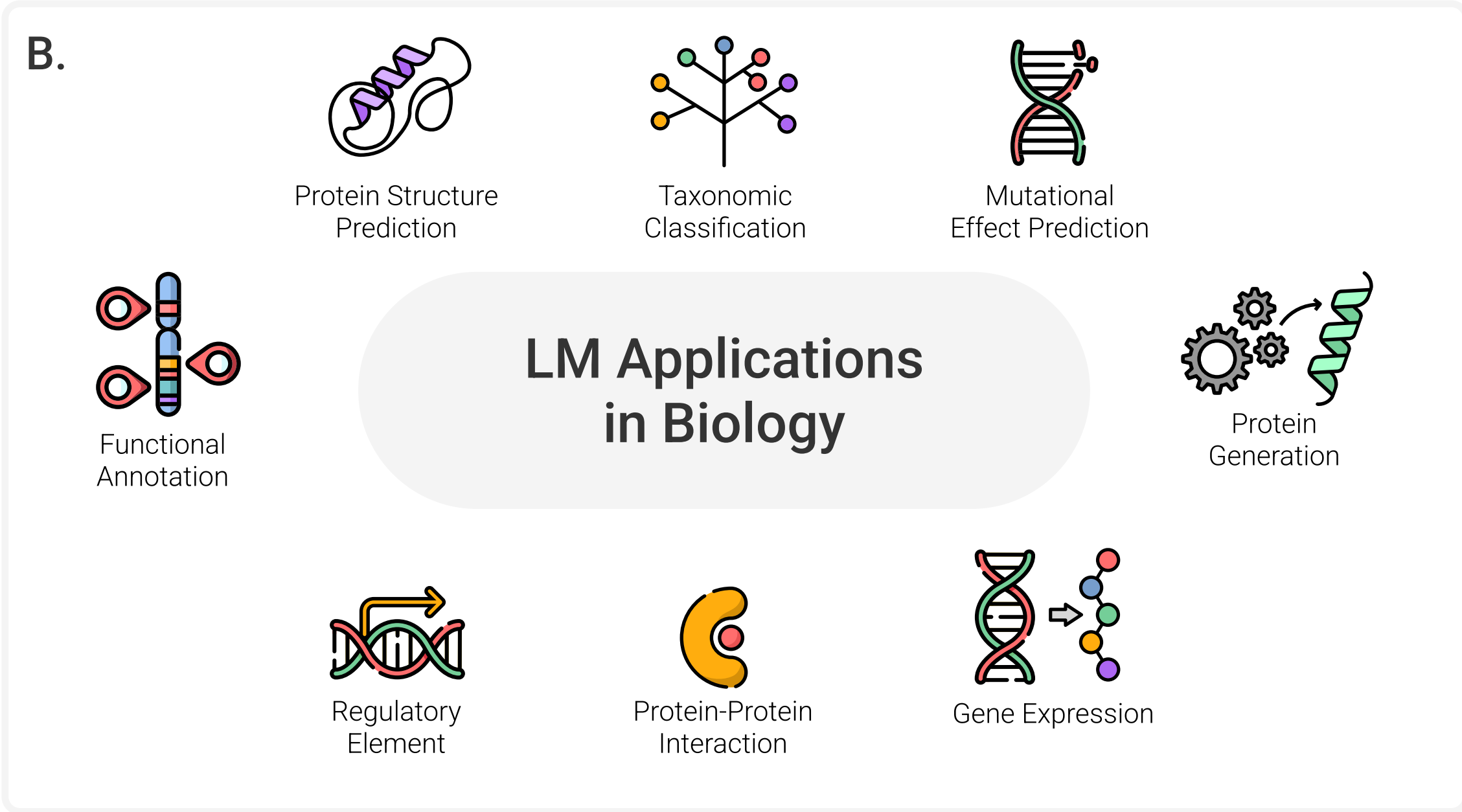

**Figure 2. Language models in biology. A.** Comparison between the English language and the different biological languages. For each language, an example of a sentence is provided in the middle column, and representative words from each language's vocabulary are displayed to the right. **B.** Examples illustrating the diverse applications of language models in biological research.

## Tokenization of Biological Languages

The application of NLP techniques to biological sequences presents unique challenges due to fundamental differences between natural languages and genetic data. Unlike most human languages, which include clear punctuation, stop words, and clearly separable structures, DNA, RNA, and protein sequences lack obvious analogs to words, sentences, and paragraphs. Consequently, it is challenging to determine whether a sub-sequence of nucleotides or amino acids constitutes a functional unit such as a domain. This lack of a clear vocabulary leads to multiple approaches for tokenizing these biological sequences (see Figure 3A).

Many algorithms use character-level tokenization, treating individual nucleotides or amino acids as tokens[36,37,44–48]. This approach allows models to provide position-specific predictions, such as the protein's secondary structure prediction, and incorporate the effects of sequence variation, such as single-nucleotide mutations that can significantly alter the protein's function. This method also maintains a very small vocabulary size and eliminates problems of sparsely represented vocabulary (where most tokens appear very infrequently). However, such tokens fail to capture broader patterns and result in long sequences that can be computationally challenging for many NLP algorithms.

Another common approach is to split sequences into constant-sized $n$-grams, known as $k$-mers[39,41,49–51]. In protein sequences, which have 20 possible characters, the number of possible $k$-mers grows exponentially with the length of $k$, often far exceeding the observed vocabulary size. This results in sparse vocabulary and the presence of many tokens not observed in the training corpus. Choosing a small $k$ mitigates this problem but results in relatively long sequences comprised of tokens that lack contextual information essential for many tasks, such as genomic analysis.

Tokenization using $k$-mers can be performed in an overlapping or non-overlapping manner. In the overlapping version, a sliding window of size $k$ and stride $t$ (commonly $t = 1$) is used to create a sequence of $k$-mers. However, this approach can lead to data leakage since each $k$-mer contains information about adjacent tokens. This affects many NLP algorithms, as most training tasks of language models are dependent on the prediction of a word based on its context (MLM, next token prediction, CBOW, etc.) or the prediction of context based on a specific word (e.g., skip-gram). The overlap between tokens can reveal information about masked tokens, thus reducing the model's ability to learn underlying patterns and generalize. This overlap also reduces efficiency as more tokens are needed to represent the sequence. Non-overlapping $k$-mers avoid some of these issues but can result in dramatic changes in tokenized sequences with the addition or deletion of a single character. Further, assuming that all significant sequence parts have the same length is unrealistic. Therefore more flexible tokenization approaches were explored.

Examples of such methods include Locality-Sensitive Hashing (LSH)[52], used to address the limitations of non-overlapping $k$-mers[53], and sub-word tokenization algorithms like WordPiece[54], SentencePiece[55], and Byte Pair Encoding (BPE)[56]. These methods start with individual characters and iteratively merge the most frequent or likely token combinations into the vocabulary, which thus captures commonly occurring sub-word units (Figure 3B). These algorithms treat each input as a raw stream without pre-tokenization assumptions, making them suitable for genome sequences where traditional word and sentence structures are absent. Sub-word tokenization creates tokens of variable length, making the training prediction tasks more challenging and potentially enhancing the model's performance. Furthermore, they effectively manage sparse vocabulary issues and handle unseen tokens by initializing the vocabulary from every possible single character.

Sub-word tokenization methods have been employed in several biological language models[40,57–60] and their effectiveness has been compared across vocabulary sizes[61], reduced amino acid alphabets[62], and different tasks[63,64]. These approaches offer the additional benefit of potentially revealing common motifs of biological importance. However, tokenization schemes with variable-length vocabularies may lead to inconsistencies, such as producing sequences of different lengths for proteins that differ by a single substitution.

Since each tokenization method presents distinct advantages and limitations, it is vital to select the appropriate tokenization approach based on the specific requirements of the task at hand before commencing training. This decision can significantly impact the model's ability to capture relevant biological information and generalize effectively.

In this review, we focus on three types of sequences that are considered "sentences" by language models: DNA/RNA sequences, protein sequences, and genomes. By examining how NLP techniques are applied to these different biological "texts," we aim to provide a comprehensive overview of the current state and potential of language model applications in bioinformatics.

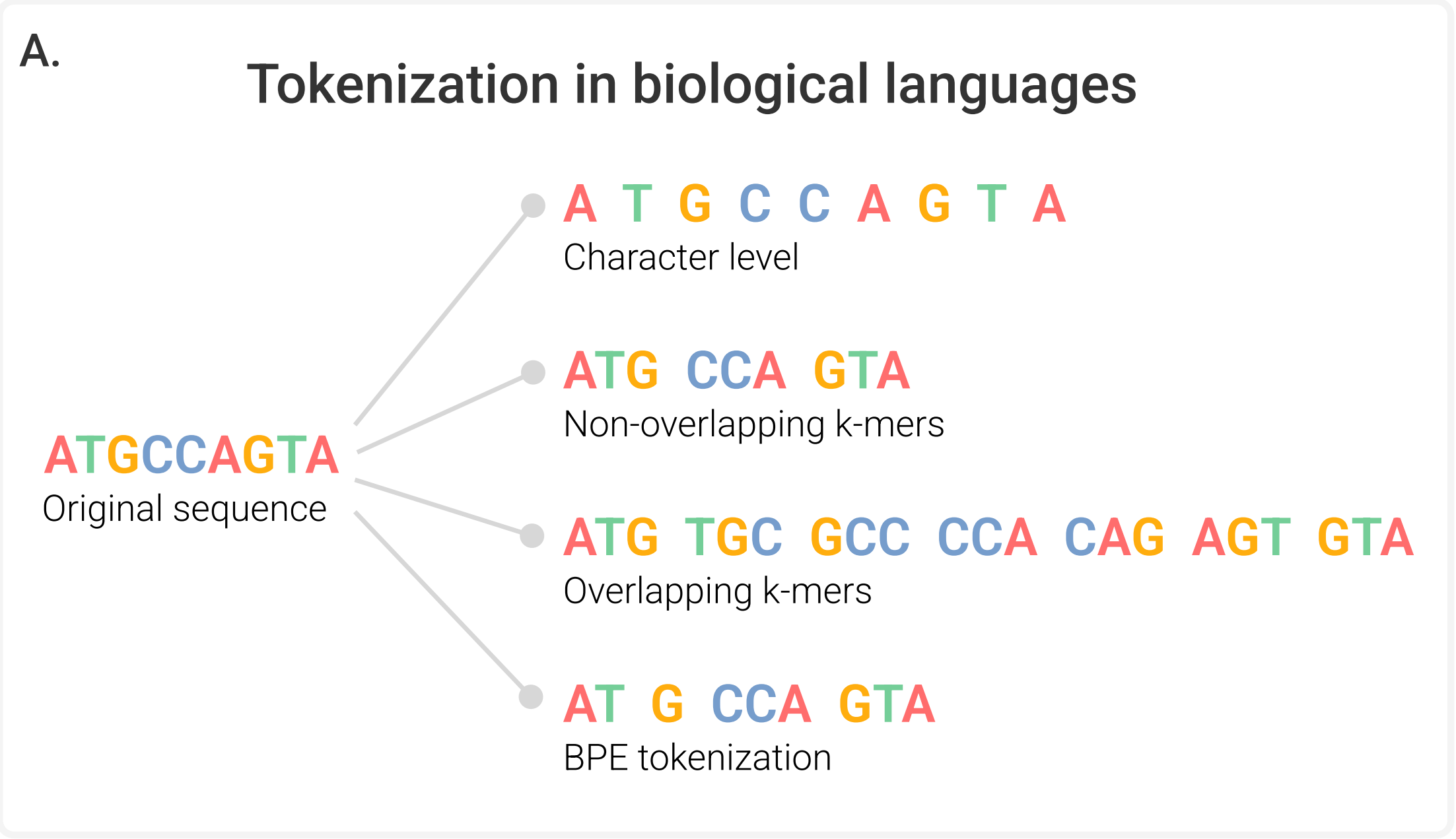

B.

**BPE Corpus Construction**

| Step | Corpus | Vocabulary |
| --- | --- | --- |
| 1 | AATTCGAAAACG | A, C, G, T |
| 2 | AA T T C G AA AA C G | A, C, G, T, AA |
| 3 | AA T T CG AA AA CG | A, C, G, T, AA, CG |
| ... | ... | ... |

**Figure 3. Tokenization of biological sequences. A.** The different tokenization methods for biological sequences. **B.** The Byte-Pair Encoding (BPE) tokenization process.

## DNA and RNA Language Models

There are numerous advantages to using DNA or RNA language models for genomic tasks. For instance, the dependency patterns encoded in DNA sequences are crucial for characterizing regulatory elements, processing non-coding sequences, and assessing the impact of individual variants[41]. Designing and training models that can capture these patterns would be extremely valuable. Moreover, such language models can also be applied to study viral genomes, which often encompass frameshift sites leading to differential translation, introducing ambiguity at the protein level[65]. Notably, RNA sequences provide additional information on the transcription level, making RNA LMs preferable when such data is relevant. Thus, language models based on RNA and DNA are widely used for numerous sequence analysis applications (Table 1).

Early approaches utilized classical NLP methods for studying DNA and RNA sequences. For example, glvdna[66] used GloVe to represent DNA *k*-mers of variable *k* in a single space for taxonomic binning, while Dna2vec[49] utilized word2vec for the representation of variable-length DNA *k*-mers, such that the cosine similarity of two vectors is highly correlated with their global sequence alignment score. Word2vec embeddings have also been shown to represent the taxonomic and metagenomic sample origin of 16S rRNA sequences[67], and have been employed to predict regulatory regions[68] and N6-methyladenosine (i.e., methylation of the adenosine nucleotide at the nitrogen-6 position) sites on mRNA sequences[69]. Further advancements in DNA models include modifications of the fastText algorithm, like fastDNA[70], which introduced noise to the training set to mimic sequencing errors, improving taxonomic binning performance, or LSHvec[53], which leveraged LSH to bin similar *k*-mers together, handling sequencing errors and large vocabulary sizes. This method demonstrated accurate taxonomic classification, even on reads with high rates of sequencing errors. More advanced methods were introduced as well, such as the Genomic Pre-trained Network (GPN)[44], which pre-trained a CNN model for genome-wide variant effect prediction, and an LSTM language model trained to predict from which lab a given DNA sequence originated[71].

The remarkable success of transformer models in natural language processing has inspired numerous applications to DNA sequences. A prominent example is DNABERT[39], a BERT-based model pre-trained on the entire human reference genome, taking as input overlapping nucleotide *k*-mers (with separate models for each *k* between 3 and 6). It has demonstrated performance comparable to the state-of-the-art for prediction of promoter regions, transcription factor binding sites (TFBSs), and splice sites. DNABERT-2[40] further improved this model by training on various organisms and changing the tokenization to BPE, thus decreasing the sequence length and mitigating the leakage problem of overlapping *k*-mers. DNABERT-2 also introduced architectural changes to overcome input length constraints and reduce the time and memory complexity. As part of this study, the Genome Understanding Evaluation (GUE), which consists of datasets and tasks for pre-training and benchmarking, was introduced. DNABERT-S[72] is a fine-tuned DNABERT-2 model optimized for taxonomic classification using the Curriculum Contrastive Learning ($C^2LR$) strategy and Manifold Instance Mixup (MI-Mix), a contrastive objective that mixes the hidden states of DNA sequences at randomly selected layers.

Another notable DNA language model is the Nucleotide Transformer[41], pre-trained on the human reference genome and 850 genomes of diverse bacterial and eukaryotic species. The model's embeddings were used to solve a diverse set of 18 genomic prediction tasks, including the prediction of epigenetic marks, promoters, and splice sites, either by fine-tuning the model or by using the embeddings as an input for another machine learning model. The Nucleotide Transformer reported superior mean performance across all of these tasks in comparison to other DNA language models, such as DNABERT[39], HyenaDNA[46], and Enformer[43] (see below). An additional DNA language model is GROVER[60], a BERT-based model trained on the human genomes using a BPE tokenization, which exhibited superior performance on some of the tasks introduced by the Nucleotide Transformer.

In contrast, the Genome-Scale Language Model (GenSLM)[73] was exclusively trained on bacterial and viral gene sequences, leveraging a hierarchical LLM. It is comprised of a generative pre-trained transformer to capture local interactions, and a diffusion-based model to capture longer-range interactions, employing genome-scale data to model individual mutations at the nucleotide level. Subsequently, GenSLM was used to generate new SARS-CoV-2 sequences, with the goal of predicting novel variants of concern. Similarly, the ProkBERT[74] model family was trained on microbial data and had shown high performance on the promoter prediction and phage identification tasks.

Specialized DNA and RNA language models were developed for specific tasks, such as prokaryotic genome annotation and transcription start sites (TSSs) identification[45], RNA N7-methylguanosine sites detection[75], prediction of pre-miRNAs in genome-wide data[76], DNA enhancers identification[77], sequence correction[78], and optimization of mRNA sequences for expression level and stability[79]. Additionally, a model was pre-trained on 5′ UTRs and supervised information such as secondary structure and minimum free energy to predict mean ribosome loading, translation efficiency, and mRNA expression level[80]. A different transformer model, BetaAlign[81], was developed for sequence alignment by converting it into a translation problem. GeneBERT[82], a multi-modal model, was trained for predicting interactions between regulatory elements using three pre-training tasks: sequence pre-training, region pre-training, and sequence region matching. It was then used for promoter classification, TFBS prediction, disease risk estimation, and splicing site prediction.

### *DNA-Level Genomic Language Models*

To overcome the length limitations of traditional transformers and process long DNA sequences, innovative architectures have emerged. Enformer[43] integrates dilated CNN with transformers to predict gene expression, chromatin states, enhancer-promoter interactions, and variant effects on gene expression, greatly increasing the network's receptive field to 100 kb to model the effects of distal enhancers and insulators on gene expression. HyenaDNA[46], Evo[47], and Evo 2[48], all based on the Hyena architecture, were trained using the next-token prediction task and can process long sequences at single-nucleotide resolution. HyenaDNA was pre-trained on the human genome using a context length of up to a million tokens, offering long-range capabilities that were shown to benefit multiple downstream tasks, surpassing the performance of DNABERT and the Nucleotide Transformer on most of them[46]. Evo is based on the stripedHyena model, which hybridizes attention and hyena operators, and uses a context length of 131 k tokens. The model exhibits impressive zero-shot prediction abilities, showcasing enhanced gene essentiality prediction as well as mRNA and protein expression level prediction comparable to other DNA LMs such as the Nucleotide Transformer and GenSLM. It also offers improved performance for zero-shot prediction of fitness of mutants compared to DNA and RNA LMs and competitive performance to protein LMs. Moreover, Evo can generate coding-rich sequences up to 650 kb in length, as demonstrated by generating synthetic CRISPR-Cas complexes and entire transposable systems[47]. Evo has demonstrated the capability to perform function-guided gene design using DNA prompts as a starting point. The generated de-novo genes, lacking significant homology to known proteins, have exhibited strong functional activity in experimental assays. Additionally, they introduced SynGenome, a database of genomic sequences generated by Evo[83]. Evo 2 expands beyond its predecessor's focus on prokaryotic genomes to encompass a comprehensive dataset spanning eukaryotic organisms as well, including human, plant, and other unicellular and multicellular species. Based on the StripedHyena 2 architecture, which integrates three distinct input-dependent convolution operators with the attention mechanism, Evo 2 is capable of processing genomic sequences extending up to one million nucleotides in length. The model exhibits strong predictive capabilities for assessing variant impact on pathogenicity and splicing, achieving state-of-the-art performance on these tasks[48].

While DNA and RNA language models demonstrate great merit for genomic analysis, they have several limitations. First, for coding regions, nucleotide sequences are longer and less information-dense than amino-acid sequences, necessitating more advanced architectures compared to protein LMs (pLMs). Second, since a single amino acid can be encoded from different codons, protein evolution operates primarily at the amino acid level, making homology better measured through protein sequences. Moreover, the protein's structure and function are more directly represented by the amino acid sequence. Therefore, pLMs are more suitable for tasks such as homology search, functional annotation, and prediction of protein structure and protein-protein interactions (PPIs).

## Protein Language Models

Protein language models have emerged as powerful tools in genomic research, demonstrating their utility across a wide range of tasks (Table 2). Early models leveraged word2vec, with ProtVec[51] being the first to apply this approach to protein sequences for tasks such as protein family classification and disordered protein prediction. ProtVec was later adapted for additional tasks such as predicting kinase activity[84], gene function[85], and MHC-epitope binding probabilities[86]. Subsequent extensions of ProtVec have demonstrated improved performance, such as Seq2vec[50], which employs doc2vec for whole-sequence embedding, and ProtVecX[57], which utilizes BPE tokenization and combines the word embeddings with *k*-mer frequencies. Other models employed word2vec for the prediction of PPIs[58], human leukocyte antigen (HLA) class I-peptide binding[87], and classification of biosynthetic gene clusters (BGCs) using Pfam domains as tokens rather than amino acids[88]. The fastText algorithm has also been adapted for protein sequences, such as for substrate identification of membrane transport proteins[89].

LSTM-based models marked the next phase in pLM development. For example, SeqVec[90], which employs ELMo for amino acid representation, has been used to predict protein secondary structure, disorder, subcellular localization, and membrane binding, as well as Gene Ontology (GO) annotation[91]. UniRep[34], a model based on multiplicative LSTM, has been shown to capture biophysical properties, phylogenetics, and secondary structures of proteins, and to predict protein stability. Another model is Multi-Task LSTM (MT-LSTM)[13], which was trained on three learning tasks simultaneously: the MLM task, residue-residue contact prediction, and structural similarity prediction. MT-LSTM was used to detect transmembrane regions and predict mutational effects. Another study trained an LSTM model to encode structural information using a two-part feedback mechanism incorporating information from structural similarity between proteins and pairwise residue contact maps for individual proteins[92], which was later used to predict PPIs[93]. Other LSTM-based models were used for enzyme classification, GO annotation, remote homology detection[94], MHC binding prediction[95], mutational effects on PPIs[96], and prediction of viral escape mutations[97].

The advent of transformer models in NLP has spurred the development of various transformer-based pLMs. Notably, ProtTrans[35] introduced a suite of transformer-based models, including two auto-regressive models (ProtTXL and ProtXLNet) and four auto-encoder models (ProtBert, ProtAlbert, ProtElectra, and ProtT5). These models have been evaluated on diverse tasks such as prediction of protein secondary structure, sub-cellular location, and protein solubility, and have since been adapted to different tasks including bacterial type IV secreted effectors prediction[98], membrane protein type classification[99], distant homology search[100,101], and structure–structure similarity search and structural alignment directly from the sequence[102]

The TAPE-Transformer[38], pre-trained on ~30 million Pfam domains, was evaluated on the TAPE set of diverse tasks, which includes secondary structure prediction, residue contact prediction, remote homology detection, and protein fitness prediction[103]. The ProteinBert[104] model employed a unique global attention approach to adapt to long sequences and was pre-trained using a dual task of MLM and GO annotation prediction to capture both local and global representations of proteins. The SaProt[105] model was trained using a structure-aware vocabulary that combines residue and 3D geometric information. Also noteworthy is RITA[106], a suite of autoregressive generative models, which had good performance on fitness effect and enzyme function prediction.

The ESM models are large transformers trained on hundreds of millions of proteins and represent some of the most prominent and comprehensive pLMs. ESM-1b[36] was shown to produce embeddings useful for remote homology detection, prediction of secondary structure, long-range residue-residue contacts, and mutational effects. ESM-1b's embeddings were also used for homology search[107], enzyme function prediction[108], prediction of bacterial type IV secreted effectors[109], membrane protein type classification[99], as well as BGC detection and characterization[110]. The MSA Transformer[111], which takes an MSA as input and utilizes both row and column attention, has outperformed ESM-1b, TAPE, and ProTrans-T5 in residue-residue contact prediction. It was also used to predict the secretion signal of bacterial type III secretion systems[112]. ESM-2[37] improved upon ESM-1b with architectural refinements and increased data and computational resources. With ESM-2, the authors also introduced ESMFold, a sequence-to-structure predictor that is nearly as accurate as alignment-based methods and is considerably faster, along with the ESM Metagenomic Atlas, a database of millions of metagenomic protein structures predicted by ESMFold. ESM2 was further used for de novo generation of proteins exhibiting low sequence similarity to naturally occurring proteins[113]. In addition, ESM-1v[114] was developed for mutational effect prediction, exhibiting a comparable performance to state-of-the-art mutational effect predictors. The embeddings of ESM-1v were also used for variant effect prediction of unseen human proteins[115]. Additional advancements include ESM3[116] and ESMC[117]. ESM3 is a multimodal generative language model trained on billions of proteins. ESM3 processes the sequence, structure, and function of proteins and was used, among other tasks, to generate a new green fluorescent protein (GFP). ESMC, which succeeds ESM2, was trained with greater computational resources on a substantially larger protein dataset that incorporates metagenomic sequences[117], and ESMFold2, which leverages ESMC embeddings to predict the structures of individual protein chains, protein complexes, and antibody–antigen complexes.

Some tasks benefit from structural information and, as such, will be more suitable for pLMs that integrate such information. For example, ProstT5[118], a model based on a ProtT5 that was further fine-tuned for the translation between amino acid sequences and the 3Di alphabet, which represents the 3D protein structure in a simple 1D sequence, has shown improved performance on structure-related tasks. Similarly, SaProt[105] is trained using a structure-aware vocabulary that combines residue and 3D geometric information. Additional such models include SeqDance and ESMDance[119], which are trained on protein sequence and structure ensembles to capture protein dynamic interactions, and MULAN[120], which is based on ESM2 with the addition of a structure adapter that integrates structural information represented by residue angles.

In addition to these general-purpose models, numerous transformer LMs have been developed for specific tasks, such as the prediction of protein structure[121,122], protein fitness[123], epitopes[124], T-cell receptor-antigen binding[125], bitter peptides[126], and lysine crotonylation sites[127], as well as generation of proteins[59,128,129] and signal peptides[130]. Notably, TALE[131] is a transformer that embeds both protein sequences and hierarchical GO labels for protein annotation, and the Co-evolution Transformer[132] incorporates homologous sequence information in MSAs for protein contact prediction. Additionally, DEDAL[133] is a model for pairwise sequence alignment that utilizes a transformer model to produce, given a pair of sequences, scoring matrices for the Smith-Waterman local alignment algorithm.

While pLMs are powerful models for a variety of genomic tasks, they have some limitations that should be addressed. First, protein sequences are more complex than DNA sequences, comprising 20 different amino acids as opposed to the four nucleotides in DNA, leading to token sparsity: $20^k$ possible tokens of length k, instead of $4^k$ for DNA sequences. This necessitates careful consideration of tokenization methods for pLMs. Secondly, pLMs do not inherently capture codon usage bias, which encodes additional information regarding the protein structure[134,135] and evolution[136,137]. To address this limitation, the cdsBERT[138] model incorporates codon information into the pre-trained ProtBert, while the CaLM[139], codonGPT[140], and SynCodonLM[141] models are pre-trained directly on the codon alphabet, and the codonTransformer[142] is trained on coding sequences represented both as nucleotides and amino acids by predicting the codon that encodes the relevant amino acid. These approaches highlight the potential for integrating codon bias into pLMs. Additionally, by focusing on individual genes or proteins, pLMs may overlook the broader genomic context of sequences. Addressing these challenges is key to the continued advancement of pLMs in genomic research.

## Gene-level Genome Language Models

Gene-level genomic language models represent an innovative approach to understanding the organization and function of genetic information, particularly in prokaryotes. These organisms organize their genomic information into gene clusters (i.e., syntenic sets of functionally related genes), and contain operons, sets of co-transcribed genes[143]. The conservation of some gene clusters across phylogenetically distant bacterial genomes, due in part to horizontal gene transfer[144], makes genomic context a powerful predictor of gene function in prokaryotes[29–31,143,144]. Therefore, LMs that leverage genomic context information could be effectively applied to these tasks across a wide range of organisms (Table 3). In this approach, instead of treating a single gene or protein as a sentence, genomic language models consider a genome, or part of it, as such.

One notable NLP approach utilized word2vec to represent gene family annotations of proteins, effectively learning the "language" of genomic context. The resulting embeddings proved useful in predicting functional annotation of uncharacterized proteins and identifying putative systems associated with microbial interaction and defence[145]. GeNLP[146], a web tool to visualize and explore more than half a million gene families in this embedding space, was also developed. A recent important model on the genome level, gLM[42], employs a transformer architecture to learn contextualized protein embeddings that capture both the protein sequences and their genomic context. The model processes 15 to 30 consecutive genes, each represented by a feature vector comprising the PCA principal components of protein embeddings generated by ESM2, concatenated with an orientation feature indicating strand direction. Pre-trained on an MLM task using these feature vectors, gLM demonstrated enhanced performance in enzyme classification compared to standard pLM embeddings, and its attention patterns were utilized for operon prediction. A recent advancement is gLM2[147], a mixed-modality genomic language model that captures contextualized representations of genomic elements. The gLM2 model is trained on the Open MetaGenomic (OMG) corpus, in which the coding sequences are represented by translated amino acids and intergenic sequences by nucleic acids. This approach enables gLM2 to perform tasks related to both protein and DNA sequences, achieving enhanced performance on protein-related tasks compared to ESM2 and comparable results to the Nucleotide Transformer on DNA-based tasks. Similarly to gLM, PlasGO[148] leverages both global and local contexts for functional annotation of plasmid-encoded proteins. The local context within protein sequences is captured using the pre-trained pLM ProtT5-XL-U50, while the global context across proteins is modelled using a BERT-based framework. This hierarchical design has been effectively applied to annotate plasmid proteins with GO terms. Similarly, Bacformer[149] learns contextualized protein embeddings by representing metagenome-assembled genomes (MAGs) as a series of proteins, each represented using mean ESM2 embeddings. The model was then used to predict PPI and operons, determine gene essentiality, and infer phenotypic traits.

Another innovative approach employed a transformer model trained on gene annotation bag-of-words representations of bacterial genomes. This model derived meaningful whole-genome representations that captured complex biological traits such as host specificity and serotypes (i.e., surface antigen variation within a species). These embeddings were also successfully used to predict the minimum inhibitory concentration (MIC) of 12 common antibiotics[150]. Additionally, the Genomic Impact Transformer (GIT)[151] was developed to detect tumour-driver somatic genomic alterations (SGAs) and predict cancer phenotypes based on its SGAs. GIT's attention mechanism integrates multiple gene embeddings of SGAs and cancer-type embeddings into a tumour embedding vector, which is then used to predict differentially expressed genes in tumours.

In conclusion, genomic language models present a promising avenue for improving our understanding of evolution, interaction, and associations between genes. As this field progresses, the development and adaptation of more sophisticated genomic LMs to diverse biological tasks could potentially unlock new insights into genomic organization and function across different organisms. Future research focusing on the refinement of these models and exploration of potential applications could prove highly beneficial for various areas of genomics and systems biology.

***Table 1. Details and descriptions of the different DNA and RNA language models mentioned in this review*.**

[a]Linking to the repository if available. OL – overlapping, NOL – Non-overlapping, nuc – nucleotide, aux – auxiliary, OHE – one-hot encoding.

| Name[a] | Language | Architecture | Tokens | Description | Ref |
|---|---|---|---|---|---|
| **NA (Mejía-Guerra, 2019)** | DNA | Word2vec | OL *k*-mers | Model for regulatory region prediction at the *k*-mer level. Uses "bag-of-*k*-mers" weighted based on TF-IDF and word2vec embeddings of *k*-mers. | [68] |
| **Gene2vec** | RNA | Word2vec | OL 3-mers | Integrates word2vec embeddings with other RNA representations as input for a CNN model for the prediction of N6-methyladenosine sites. | [69] |
| **dna2vec** | DNA | Word2vec | NOL *k*-mers | Variable-length *k*-mers embedding, achieving vector addition analogous to nucleotide concatenation and correlation with Needleman-Wunsch scores. | [49] |
| **fastDNA** | DNA | FastText | OL *k*-mers | Modified fastText for *k*-mer embedding, includes random mutations and reverse-complement reads to address sequencing errors. | [70] |
| **LSHvec** | DNA | FastText, LSH | NOL *k*-mers | Uses LSH to bin similar *k*-mers together, thus addressing the problems of vocabulary size and sequencing errors. | [53] |
| **glvdna** | DNA | GloVe | OL *k*-mers | GloVe-based model for representing *k*-mers from variable *k* in one space. | [66] |
| **GPN** | DNA | CNN | Single nuc | Predicts genetic variant effects using a convolutional neural network with a masked language model task on a single nucleotide level. | [44] |
| **NA (Alley, 2020)** | DNA | LSTM | BPE | Model for genetic engineering attribution prediction using LSTM sequence embeddings and phenotypic metadata information. | [71] |
| **miRe2e** | RNA | Transformer | Nuc OHE | Transformer model for predicting pre-miRNAs in genome-wide data. | [76] |
| **DNA-transformer** | DNA | Transformer | Single nuc | Prokaryotic genome annotation model for identifying transcription start sites in *Escherichia coli*. | [45] |
| **Bert-Enhancer** | DNA | Transformer | Single nuc | Uses BERT and 2D CNN to identify DNA enhancers from sequence information. | [77] |
| **UTR-LM** | RNA | Transformer | Single nuc | Pre-trained on 5′ UTRs from multiple species and incorporates secondary structure and minimum free energy information. Fine-tuned for tasks like ribosome loading and mRNA expression level prediction. | [80] |
| **BetaAlign** | DNA | Transformer | Single nuc | Aligns novel sets of sequences by converting the alignment problem into a sequence-to-sequence learning problem. | [81] |
| **DeepConsensus** | DNA | Transformer | Single nuc + aux features | Gap-aware transformer-encoder for sequence correction, read error detection, and improvement of genome assembly and variant calling. | [78] |
| **GeneBERT** | DNA | Transformer | OL *k*-mers | Pre-trained on large-scale genomic data for predicting regulatory elements interactions, using multi-modal and self-supervised tasks. | [82] |
| **DNABERT** | DNA | Transformer | OL *k*-mers | Pre-trained on the human genome and fine-tuned for tasks like predicting promoter regions and transcription factor binding sites. | [39] |
| **DNABERT 2** | DNA | Transformer | BPE | Improved version of DNABERT by replacing the tokenizer, updating the architecture for efficiency, and training on multiple organisms. | [40] |
| **ProkBERT** | DNA | Transformer | OL *k*-mers | Family of transformer models trained on bacteria, archaea, viruses, and fungi using Local Context-Aware (LCA) tokenization, where individual elements consist of overlapping k-mers. Finetuned on tasks of promoter prediction and phage sequence identification. | [74] |
| **The Nucleotide Transformer** | DNA | Transformer | NOL 6-mers | Pre-trained on diverse human genomes and other species, used for various genomic prediction tasks. | [41] |
| **CodonBERT** | RNA | Transformer | NOL 3-mers | mRNA LLM for optimal sequence selection for specific proteins or peptides, focusing on mRNA vaccines. | [79] |
| **CaLM** | DNA | Transformer | NOL 3-mer | Codon bases LM pre-trained on cDNA sequences. Evaluated on the tasks of taxonomic classification, melting point prediction, solubility prediction, subcellular localization prediction, and function prediction, and used to predict transcript and protein abundance. | [139] |
| **codonGPT** | RNA | Transformer | NOL 3-mer | A generative LM for protein-coding mRNA trained on sequences from model organisms. It uses reinforcement learning to optimize the codons of generated sequences according to user requirements such as CAI, GC content, RNA stability, codon diversity, and repeat usage. | [140] |
| **SynCodonLM** | RNA | Transformer | NOL 3-mer | A codon LM that outperforms prior models on benchmarks sensitive to DNA-level features, including mRNA and protein expression. | [141] |
| **Grover** | DNA | Transformer | BPE | BERT-based human genome model, evaluated on promotor classification and prediction of protein–DNA binding, splice sites, and enhancers. | [60] |
| **Enformer** | DNA | Transformer, CNN | Nuc OHE | Combines dilated CNN and transformer architecture to predict gene expression and chromatin states from DNA sequences, increasing receptive field from 20 to 100 kb to model distal enhancers. | [43] |
| **BERT-m7G** | RNA | Transformer, Stacking Ensemble | Single nuc | Detects RNA N7-methylguanosine sites using a stacking ensemble classifier that utilizes features selected from BERT embeddings using Elastic Net. | [75] |
| **GenSLMs** | DNA | Transformer, Diffusion Model | NOL 3-mers | Pre-trained on bacterial and viral gene sequences, uses hierarchical modeling to capture short and long-range interactions. Fine-tuned to generate new viral sequences. | [73] |
| **HyenaDNA** | DNA | Hyena hierarchy | Single nuc | LLM pre-trained using next-token predictions on human genome sequences. Offers subquadratic time complexity and can process up to a million nucleotides. | [46] |
| **Evo** | DNA | StripedHyena | Single nuc | Models long DNA sequences using a context length of 131 kbps. Trained on prokaryotic and phage genomes and used to predict mutant fitness, gene function, gene essentiality, and generate coding sequences up to 650 kb in length. | [47] |
| **Evo 2** | DNA | StripedHyena 2 | Single nuc | Expands upon Evo by incorporating eukaryotic genomes into its training corpus and can process sequences up to 1 million nucleotides. Evo2 predicts mutational effects on proteins, RNA, and organismal fitness and enables genome-scale generation across all domains of life. | [48] |

**Table 2. Details and descriptions of the different protein language models mentioned in this review**.
[a]Linking to the repository if available. OL – overlapping, NOL – Non-overlapping, aa - Amino Acid.

| Name[a] | Architecture | Tokens | Description | Ref |
|---|---|---|---|---|
| **HLA-vec** | Word2vec | Single aa | Word2vec embeddings of peptides are used as inputs for a CNN to predict HLA class I-peptide binding. | [87] |
| **ProtVec** | Word2vec | 3-mers | Uses skip-gram word2vec to extract vector representations. Used for protein family classification and disordered protein prediction. | [51] |
| **ProtVecX** | Word2vec | BPE | Extends ProtVec by combining *k*-mer occurrence with embeddings, thus improving sequence classification performance. | [57] |
| **Bio2Vec** | Word2vec | SentencePiece | Bio2Vec trained a skip-gram word2vec model and used the embeddings as input for a CNN to predict protein-protein interactions. | [58] |
| **DeepBGC** | Word2vec | Pfam domains | Classifies BGCs and their products using BiLSTM on word2vec-like embeddings of Pfam domains. | [88] |
| **Seq2vec** | doc2vec | OL+NOL *k*-mers | Embeds entire protein sequences as documents using doc2vec. It showed better performance than ProtVec in protein family classification. | [50] |
| **FastTrans** | FastText | OL *k*-mers | Uses fastText skip-gram to create embeddings for predicting substrate specificities of membrane transport proteins. | [89] |
| **MuPIPR** | LSTM, CNN | Single aa | Predicts mutation effects on PPIs by propagating point mutation effects. Uses a siamese recurrent CNN to encode wild-type and mutant protein pairs. | [96] |
| **UniRep** | mLSTM | Single aa | A model for embedding protein sequences. It predicts the structural and functional properties of proteins and can be used for protein design. | [34] |
| **SeqVec** | BiLSTM | Single aa | A model for amino acid representation for whole protein sequence. Predicts secondary structure, disorder localization, and membrane binding. | [90] |
| **NA (Bepler, 2019)** | BiLSTM | Single aa | Trained using a two-part feedback mechanism incorporating global structural similarity and pairwise residue contact maps. | [92] |
| **UDSMProt** | BiLSTM | Single aa | Model for remote homology detection and functional annotations of proteins. | [94] |
| **MT-LSTM** | BiLSTM | Single aa | Trained on MLM, residue-residue contact prediction, and structural similarity prediction. Predicts transmembrane regiona and vatiant phenotype. | [13] |
| **NA (Hie, 2021)** | BiLSTM | Single aa | Models viral escape mutations, focusing on mutations that maintain infectivity but alter immune recognition. | [97] |
| **ProtTrans** | Transformer | Single aa | Diverse transformer models trained on protein sequences. Reports accurate protein secondary structure, sub-cellular location, and more. | [35] |
| **TAPE** | Transformer | Single aa | A transformer pre-trained on ~30 million Pfam domains, evaluated on the TAPE collection of biological tasks for benchmarking protein LMs. | [38] |
| **SPGen** | Transformer | Single aa | Generates signal peptides (SPs) by "translating" a mature protein sequence without SPs to the corresponding SP sequence. | [130] |
| **ESM-1b** | Transformer | Single aa | A large pLM designed for the prediction of residue-residue contacts, remote homology, secondary structure, and mutational effect. | [36] |
| **ESM2** | Transformer | Single aa | Improved ESM-1b in terms of architecture and training parameters. Introduced ESMFold, a sequence-to-structure predictor. ESM2 and ESMFold were utilized to create the ESM Metagenomic Atlas with over 600 million metagenomic proteins. | [37] |
| **ESMC** | Transformer | Single aa | Improved ESM2 in terms of training compute and data. Introduced ESMFold2, that predicts the structure of both individual protein chains and biomolecular complexes. | [117] |
| **ESM3** | Transformer | Discrete per modality | Multimodal generative LM that takes as input the sequence, structure, and function of proteins. Reported to generate new proteins with high accuracy. | [116] |
| **ESM-1v** | Transformer | Single aa | Predicts mutational effects by scoring each possible mutation in the protein and ranking the mutant protein's relative activity. | [114] |
| **MSA Transformer** | Transformer | Single aa | The MSA Transformer is designed to capture global co-evolutionary patterns by considering information from all homologous sequences in an MSA. It is used for tasks like residue contact prediction. | [111] |
| **BERT4Bitter** | Transformer | Single aa | A model for predicting bitter peptides from the amino acid sequences. | [126] |
| **TALE** | Transformer | Single aa | A protein function annotator that embeds both protein sequences and function labels (hierarchical GO terms). | [131] |
| **EpiBERtope** | Transformer | Single aa | Transformer language model for linear and structural epitope prediction using only protein sequence as input. | [124] |
| **ReLSO** | Transformer | Single aa | Model for protein generation and fitness prediction, optimizing both simultaneously. | [129] |
| **RGN2** | Transformer | Single aa | Uses a BERT model to learn latent structural information and predicts protein structure from sequence, considerably faster than AlphaFold2. | [121] |
| **OmegaFold** | Transformer | Single aa | Predicts high-resolution protein structure from a single protein sequence. Combines a pLM, OmegaPLM, and a geometry-inspired transformer model trained on protein structures. Roughly ten times faster with comparable or better accuracy to MSA-based methods such as AlphaFold2 . | [122] |
| **BERT-Kcr** | Transformer | Single aa | Predicts lysine crotonylation post-translational modification sites using a BiLSTM network for classification. | [127] |
| **ProteinBERT** | Transformer | Single aa | Combines language modeling with gene ontology annotation prediction, capturing local and global representations of proteins. Utilizes linear global attention to decrease memory and running time complexity. | [104] |
| **RITA** | Transformer | Single aa | A suite of autoregressive generative models, evaluated on the tasks of next amino acid prediction, zero-shot fitness, and enzyme function prediction. | [106] |
| **DEDAL** | Transformer | Single aa | Performs pairwise alignment by integrating sequence embeddings and a flexible parameterization of the Smith-Waterman algorithm. | [133] |
| **TCR-BERT** | Transformer | Single aa | An LM trained on T-cell receptor sequences to predict to which antigens a specific TCR will bind. | [125] |
| **ProGen** | Transformer | Single aa + tags | An LM for protein generation according to the given requirements and constraints. Uses taxonomic and keyword tags. | [128] |
| **ProstT5** | Transformer | Single aa or 3Di | Based on the ProtT5 model, it translate between protein sequence and structure, enabling accurate prediction of structural and functional properties. | [118] |
| **SaProt** | Transformer | Single aa + structure info | Utilized the ESM-2 architecture and a vocabulary that combines residue and structural information. Evaluated on the tasks of prediction of mutational effects, thermostability, PPI, metal ion binding, functional annotation, and protein localization. | [105] |
| **MULAN** | Transformer | Single aa+residue angles | A multimodal model that integrates both sequence information as ESM-2 embedding and residue angles. Exhibited improved performance on variety of tasks such as thermo-stability, Fluorescence and PPI prediction. | [120] |

| **SeqDance & ESMDance** | Transformer | Single aa+protein dynamic properties | Trained on molecular dynamics-derived biophysical properties to capture protein conformational dynamics, enabling improved prediction of mutation effects on protein folding stability. ESMDance, built upon ESM2 outputs, outperforms it in zero-shot prediction of mutation effects for proteins lacking evolutionary information. | [119] |
|---|---|---|---|---|
| **Tranception** | Transformer | NOL *k*-mers | Leverages autoregressive predictions and retrieval of homologous sequences at inference time for fitness prediction. Also introduced ProteinGym, an extensive set of deep mutational scanning assays curated to enable comparisons of various mutation effect predictors in different regimes. | [123] |
| **cdsBERT** | Transformer | NOL 3-mers | A ProtBERT-based pLM with a codon vocabulary trained on a massive corpus of CDS. | [138] |
| **CodonTransformer** | Transformer | Single aa+codon | Trained on amino acid-codon pairs from 164 organisms. Shown to generate host-specific DNA sequences with natural-like codon distribution. | [142] |
| **Co-evolution Transformer** | Transformer | MSA | Captures global co-evolutionary patterns by considering the information of the homologous sequences in an MSA while reducing the impact of non-homologous sequences. | [132] |
| **ProtGPT2** | Transformer | BPE | Language model for generating de novo protein sequences distantly related to natural ones, but with similar structural properties. | [59] |

***Table 3. Details and descriptions of the different gene-level genomic language models mentioned in this review*.**
[a]Linking to the repository if available. aa - Amino Acid, nuc – nucleotide.

| Name[a] | Architecture | Tokens | Description | Ref |
|---|---|---|---|---|
| **GeNLP** | Word2vec | Gene families | Trained on gene family annotations to learn genomic context. Used to predict functional categories and discover systems related to microbial interactions and defense. | [145] |
| **GIT** | Transformer | Gene embeddings | An encoder-decoder model to predict differentially expressed genes in tumors and detect potential cancer drivers. Combines gene embeddings of somatic genomic alteration and cancer-type embedding. | [151] |
| **NA (Naidenov, 2024)** | Transformer | BOW of gene annotations | Trained on bacterial genome assemblies to derive whole genome embeddings representing complex biological characteristics. Used the embeddings to predict serotypes, MIC, and other traits. | [150] |
| **gLM** | Transformer | Protein embedding | Learns contextualized protein embeddings capturing genomic context and protein sequence information. It uses PCA of ESM2 pLM embeddings for an MLM pretraining task. Applied to classify viruses from bacterial and archaeal sub-contigs. | [42] |
| **gLM2** | Transformer | Single aa + single nuc | Mixed-modality language model that learns contextualized representations of genomic elements. Trained on coding sequences represented by amino acids and intergenic regions represented by nucleotides. Applied on both DNA and protein tasks, and have shown potential at predicting PPI interfaces and identifying regulatory syntax and non-protein-coding elements in intergenic regions. | [147] |
| **PlasGo** | Transformer | Protein embedding | Utilizes a hierarchical framework that integrates both local and global protein context for GO term annotation of plasmid proteins. Employes ProtT5-XL-U50 pLM embeddings with a BERT-based model that learn the gene context. | [148] |
| **Bacformer** | Transformer | Protein embedding | Learns contextualized protein embeddings by representing metagenome-assembled genomes as sequences of ESM2 protein embeddings. Applied to protein–protein interaction and operon prediction, gene essentiality estimation, and phenotype inference. | [149] |

## Discussion

The application of NLP techniques to genomic sequences has demonstrated remarkable versatility and potential across a wide spectrum of biological tasks. Biological language models were successfully utilized for protein generation[59,113,116,128,129], protein structure prediction[35–38,90,121,122], taxonomic classification[53,66,67,70], mutational effect prediction[13,36,44,103,106,114,123], protein-protein interaction prediction[58,93], distant homology detection[38,94,100,101], gene expression prediction[43,47,80], regulatory elements identification[39,40,68,82], functional annotation of proteins[42,85,91,94,108,131,145,148], and more. The breadth of these applications underscores the transformative potential of NLP approaches in genomics.

Our review examined the diverse architectures employed in these language models, each with its unique strengths and limitations. The evolution of these models, from static word embedding techniques to sophisticated transformer-based architectures and beyond, reflects the rapid progress in this field. However, it also highlights the importance of careful model selection based on the specific requirements of each biological task and the nature of the data analyzed.

Classical methods for distributed representations of sequences, such as word2vec, fastText, and GloVe, are intuitive and relatively quick to train. However, they fall short in addressing long-distance dependencies and polysemy (words that can have multiple meanings depending on their setting), since they generate a single vector for each token irrespective of its context. In biological sequences, the context of a word in a sentence is often analogous to the role of a particular nucleotide or amino acid in the entire sequence. Hence, addressing polysemy in biological sequences is crucial, making these methods suboptimal. Recurrent Neural Networks (RNNs), especially Long Short-Term Memory (LSTM) networks, address the polysemy issue by generating contextual embeddings based on the hidden states of other tokens in the sequence but still fail to effectively capture long-term dependencies. Such limitations are problematic, especially for tasks requiring the analysis of distal interactions, such as structural predictions and gene interaction in trans.

The transformer architecture overcomes these limitations by processing entire sentences simultaneously and calculating token dependencies using the attention mechanism. Despite their state-of-the-art performance across various domains, transformers are constrained by their quadratic computational and memory complexity, limiting the length of sequences they can effectively process. This limitation is particularly relevant for long DNA sequences, such as those found in eukaryote genomes, which contain extensive non-coding regions. Additionally, transformers consist of a vast number of parameters and require substantial computational resources and large datasets for effective training. As a result, RNNs may be more suitable for tasks with fewer available sequences, such as specific protein datasets, or in situations where computational resources are limited. In contrast, transformers excel as general-purpose pre-trained models, learning a wide range of biological and evolutionary properties from comprehensive sequence databases. The expensive pre-training step is performed once and can later be applied to other tasks through transfer learning[152]. In general, large language models continue to demonstrate improved performance on extensive databases. However, for smaller, specialized datasets, the optimal approach often involves careful selection of model architectures and hyperparameter tuning or fine-tuning of pre-trained models.

The rapid growth of biological data and the increasing complexity of genomic analysis necessitate the development of more efficient models and architectures. Reducing the time and space complexity of transformer models will enable the processing of long sequences, such as entire DNA sequences, facilitating full-genome analysis. To address these challenges, several adaptations of the transformer architecture, alongside novel architectures, have been developed. For instance, algorithms that utilize sparse attention, such as Longformer[153], which employs dilated sliding windows and global attention strategies; Big Bird[154], which integrates random, window, and global attention; and Reformer[155], which replaces the dot-product attention with locality-sensitive hashing-based attention. Linear-attention[156]-based algorithms, such as the Linear Transformer[157], Linformer[158], and Performer[159], also show promise. In addition, FlashAttention[160–162], an IO-aware attention algorithm that reduces run time by reducing the number of memory operations has also been introduced. Additionally, novel architectures like Hyena[27], which combines long convolutions and data-controlled gating, and Mamba[163], a selective state space model[164] that offers linear scaling in sequence length, have emerged and proved effective. Some of these algorithms have been applied to biological sequences. For instance, the Hyena[46–48]architecture, as well as the integration of transformers with CNNs[43] have been used to process long DNA sequences, while global attention has been applied to process protein sequences and GO annotations[104]. Despite these advancements, standard transformers currently remain the most widely used architecture in genomic NLP studies, highlighting the need to explore more efficient algorithms for biological applications.

The abundance of foundation biological LMs[35–41,46–48,60,117] provides a set of five semi-supervised learning tasks focused on protein structure and fitness; ProteinGym[123] includes mutational effects of substitution and indel variants, The Genome Understanding Evaluation (GUE)[40] dataset encompasses 36 datasets across nine tasks and multiple species, including the detection of promoters, transcription factors, and splice sites, as well as species classification; and the PETA[64] benchmark, which spans 15 biological tasks across 33 datasets, including fitness prediction, PPI prediction, protein localization prediction, solubility prediction, and fold prediction. Despite these significant advances, some research gaps remain. Future efforts should prioritize expanding these benchmarks to address currently underrepresented biological research areas and include a more diverse range of organisms to ensure the robustness and generalizability of biological LMs across diverse contexts. It is also critical to create benchmarking datasets tailored toward genomic LMs. Continuously refining and broadening these evaluation frameworks will allow researchers to more effectively leverage the potential of foundation language models in biological research.

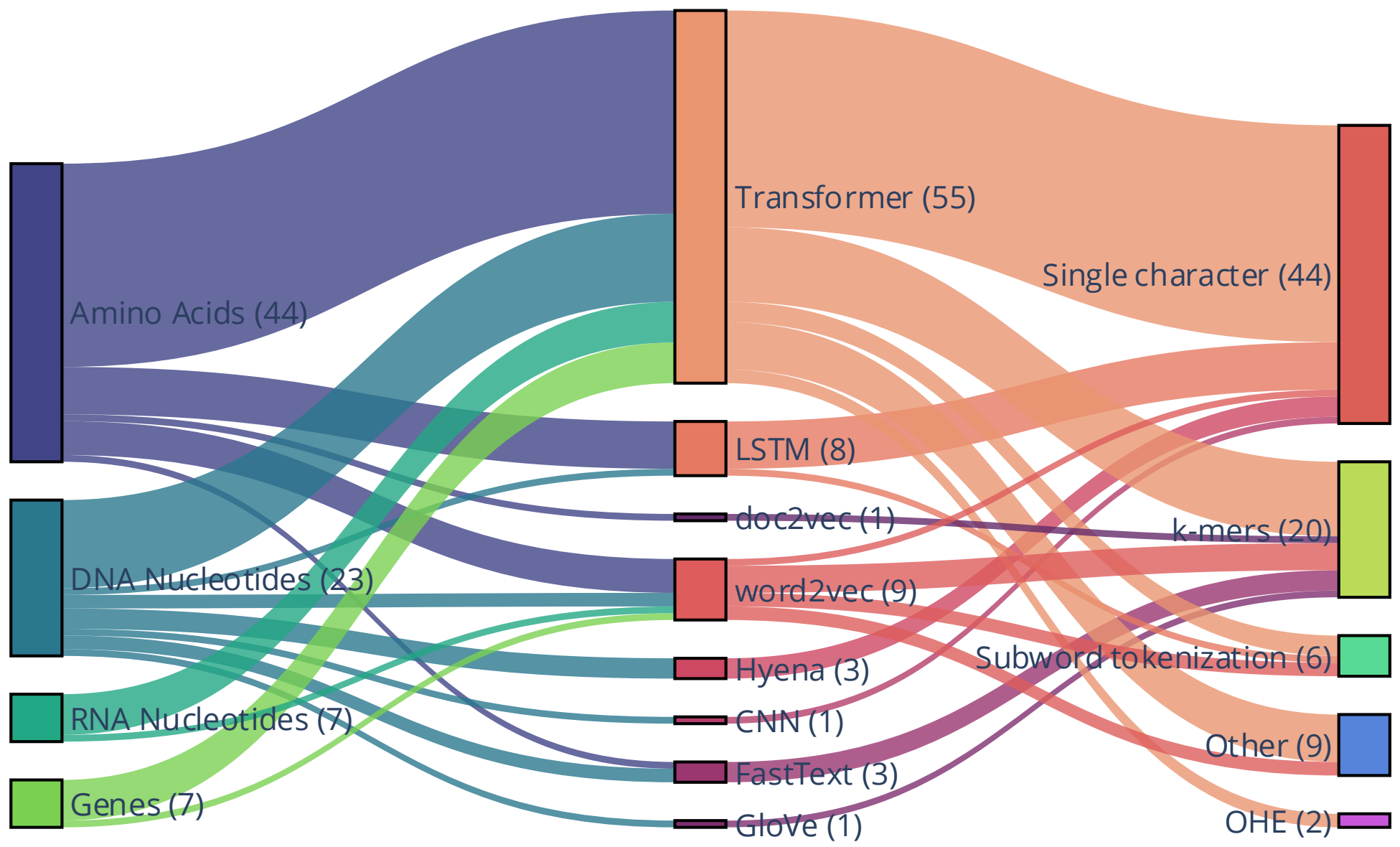

**Figure 4. Sankey diagram illustrating the primary language models (LMs) discussed in this review, categorized by the biological language they process, their architecture, and tokenization approach**. The total number of models with each attribute is indicated in brackets. OHE – One-Hot Encoding.

In addition, most biological language models treat single or groups of nucleotides or amino acids as words (See Figure 4). However, only a few models currently consider the genomic context of genes[42,145,148], which can provide crucial insights into gene function[29–31]. Further research on the study of the "gene-level genomic language" could be highly beneficial. New modeling approaches of biological sequences, such as representing proteins as a sequence of their domains, could also play an important role in developing better biological LMs[28,88,110].

There is also great potential for inferring biological conclusions from bioinformatical transformer models and their attention maps. While attention scores of protein language models have already been used to infer residue contact maps and other structural and functional conclusions[36,37,111,165,166], similar studies on DNA and genomic language models are lacking. Analysis of attention maps from these models might provide insights into splicing or transcription signals in DNA sequences, or operons and regulatory elements in genome-based languages. For example, most of the attention heads of the Transformer-XL-based DNA transformer could successfully identify and characterize transcription factors' binding sites and consensus sequences[45], demonstrating the unique potential of transformers for genome annotation and extraction of biological insights. Moreover, the confidence scores these models produce can be utilized to identify conservation patterns and differentiate between coding and non-coding regions.

In conclusion, the application of NLP techniques to biological sequences is a powerful approach, but it requires careful consideration of the specific task, data availability, and computational resources when selecting the most appropriate biological language, tokenization approach, and model architecture. While this review has highlighted the application of NLP methods to various biological tasks, many more biological questions could potentially be addressed using this approach. Specifically, classical NLP tasks could be creatively adapted to bioinformatic problems, such as applying translation problem solutions to sequence alignment[81] and signal peptide generation[130], summarization for splicing prediction, and closed-domain question answering for genomic islands detection. The field of NLP in genomics and proteomics is rapidly evolving, with significant potential for addressing complex biological questions. As researchers continue to develop more efficient and specialized models and creatively adapt NLP techniques to biological contexts, we can expect substantial advances in our ability to interpret sequence data and our understanding of biological processes.

## Acknowledgments

This research was partly funded by ISF grant number 355/23. ER and was funded in part by the Edmond J. Safra Center for Bioinformatics at Tel Aviv University. We thank Iris Burstein for the design of the figures presented in this review.

## Contributions

E.R., D.B. conceived and designed the review; E.R. executed the research and analysis; E.R., D.B. wrote the manuscript.

## Competing interests

The authors declare no competing financial or non-financial interests.